\documentclass[11pt,a4paper]{article}
\usepackage[nohyperref]{acl2019}
\usepackage{times}      % Packages for consistent fonts.
\usepackage{latexsym}   % Packages for consistent fonts.
\usepackage{url}            % simple URL typesetting
\usepackage{booktabs}       % professional-quality tables
\usepackage{amsfonts}       % blackboard math symbols
\usepackage{amsmath}
\usepackage{nicefrac}       % compact symbols for 1/2, etc.
\usepackage{microtype}      % microtypography
\usepackage{multirow}
\usepackage[OT1]{fontenc} 
\usepackage{subfig}
\usepackage[inline]{enumitem}
\usepackage{chemformula}
\usepackage{textcomp}
\usepackage{graphics} % for pdf, bitmapped graphics files
\usepackage{graphicx}
\usepackage{titling}
\usepackage{hyperref}

 \usepackage{endnotes}
 \let\svthefootnote\thefootnote
 \newcommand\freefootnote[1]{%
   \let\thefootnote\relax%
   \footnotetext{#1}%
   \let\thefootnote\svthefootnote%
 }
\setlist[description]{
  font={\ttfamily} % set the label font
}
\definecolor{darkblue}{rgb}{0, 0, 0.5}
\hypersetup{colorlinks=true,citecolor=darkblue, linkcolor=darkblue, 
urlcolor=darkblue}

\graphicspath{{./figures/}}
\providecommand{\dodraft}{false}
\newboolean{isdraft}
\setboolean{isdraft}{\dodraft}
\ifthenelse{\boolean{isdraft}}{%

\newcommand{\sheshera}[2]{{\color{red}{#1 $\to$ {\bf Sheshera}}: #2}}
\newcommand{\hawshiuan}[2]{{\color{blue}{#1 $\to$ {\bf Haw-Shiuan}}: #2}}

\newcommand{\eddie}[2]{{\color{OliveGreen}{#1 $\to$ {\bf Eddie}}: #2}}
\newcommand{\zach}[2]{{\color{magenta}{#1 $\to$ {\bf Zach}}: #2}}
\newcommand{\kevin}[2]{{\color{Bittersweet}{#1 $\to$ {\bf Kevin}}: #2}}
\newcommand{\jeff}[2]{{\color{BrickRed}{#1 $\to$ {\bf Jeff}}: #2}}
\newcommand{\elsa}[2]{{\color{green}{#1 $\to$ {\bf Elsa}}: #2}}
\newcommand{\andrew}[2]{{\color{orange}{#1 $\to$ {\bf Andrew}}: #2}}
\newcommand{\mitfolk}[2]{{\color{Maroon}{#1 $\to$ {\bf MITFolk}}: #2}}
\newcommand{\umassfolk}[2]{{\color{Maroon}{#1 $\to$ {\bf UMassFolk}}: #2}}
\newcommand{\everyone}[2]{{\color{Purple}{#1 $\to$ {\bf Everyone}}: #2}}
\newcommand{\todo}[1]{{\color{red}{{\bf TODO: #1} }}}

} {

\newcommand{\sheshera}[2]{}
\newcommand{\hawshiuan}[2]{}
\newcommand{\eddie}[2]{}
\newcommand{\zach}[2]{}
\newcommand{\kevin}[2]{}
\newcommand{\jeff}[2]{}
\newcommand{\elsa}[2]{}
\newcommand{\andrew}[2]{}
\newcommand{\mitfolk}[2]{}
\newcommand{\umassfolk}[2]{}
\newcommand{\everyone}[2]{}
\newcommand{\todo}[1]{}

}

\aclfinalcopy % Uncomment this line for the final submission
\title{The Materials Science Procedural Text Corpus: Annotating Materials 
Synthesis Procedures with Shallow Semantic Structures}

\author{{\bf Sheshera Mysore$^{1*}$ \qquad Zach Jensen$^{2*}$ \qquad Edward 
Kim$^2$ \qquad Kevin Huang$^2$}\\
  {\bf Haw-Shiuan Chang$^1$ \qquad Emma Strubell$^1$ \qquad Jeffrey 
Flanigan$^1$} \\
  {\bf Andrew McCallum$^1$ \qquad Elsa Olivetti$^2$}\\[3pt]
  $^1$College of Information and Computer Sciences\\
  University of Massachusetts Amherst\\
  \texttt{\{smysore, hschang, strubell, jflanigan, mccallum\}@cs.umass.edu}\\
  [5pt]
  $^2$Department of Materials Science and Engineering\\
  Massachusetts Institute of Technology\\
  \texttt{\{zjensen, edwardk, kjhuang, elsao\}@mit.edu}}

\date{}

\begin{document}
\maketitle

\begin{abstract}
Materials science literature contains millions of materials synthesis procedures
described in unstructured natural language text. Large-scale analysis of these 
synthesis 
procedures would facilitate deeper scientific understanding of materials 
synthesis and enable automated synthesis planning. Such analysis 
requires extracting structured representations of synthesis procedures 
from the raw text as a first step. 
To facilitate the training and evaluation of synthesis extraction models, we 
introduce a dataset of 230 synthesis procedures annotated by domain experts 
with labeled graphs that express the semantics of the synthesis sentences. The 
nodes in this graph are synthesis operations and their typed arguments, and 
labeled edges specify relations between the nodes. We describe this new resource 
in detail and highlight some specific challenges to annotating scientific text 
with shallow semantic structure. We make the corpus available to the community 
to promote further research and development of scientific information 
extraction systems.
\end{abstract}

\section{Introduction}
\label{sec-introduction}
\begin{figure}[t]
    \centering
\fbox{\includegraphics[width=0.45\textwidth]{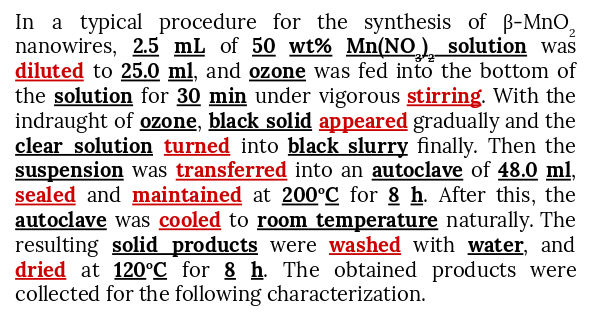}}
    \caption{Example synthesis procedure text from a materials journal article 
\citep{dong2009beta}. \textcolor{Red}{\textbf{Bold red}}
indicates the operations (predicates) involved in the synthesis; \textbf{bold 
black}
indicates arguments; \underline{underlines} demarcate entity boundaries.}
    \label{fig-text-example}
\end{figure}
\freefootnote{$^*$Equal contribution}

\begin{figure*}[t]
    \centering
\fbox{\includegraphics[width=\textwidth]{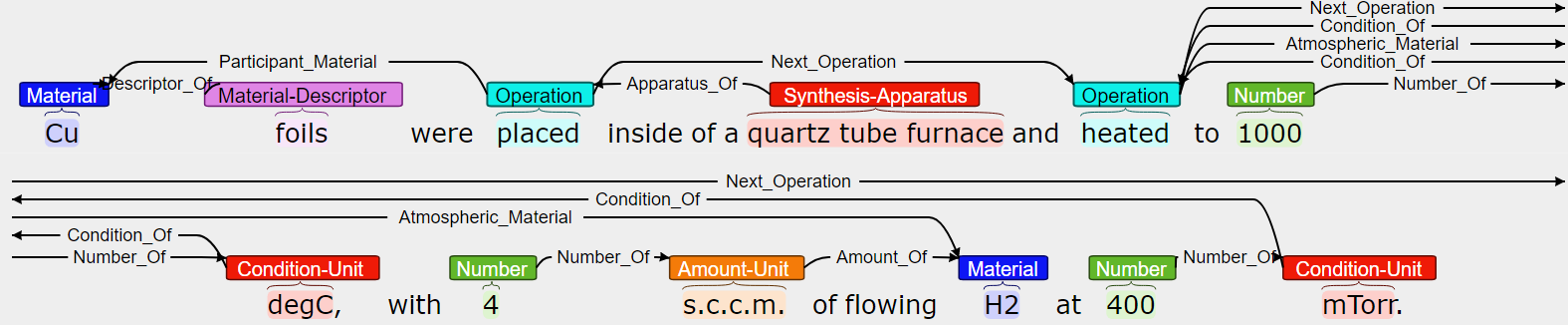}}
    \caption{An example annotated sentence. Shallow semantic 
structures generally consist of verbal predicates and arguments of these 
predicates as nodes and labeled edges between predicate and argument nodes, 
example. Heated(\textit{Condition\_of}: degC, 
\textit{Atmospheric\_Material}: 
H2, \textit{Condition\_of}: mTorr). We also label relations between argument 
entities and non-predicate entities, for example. \textit{Descriptor\_of}(Cu, 
foils) and relations between predicates, for example. 
\textit{Next\_Operation}(placed, heated).}
    \label{fig-ann-example}
\end{figure*}
Systematically reducing the time and effort required to synthesize novel 
materials remains one of the grand challenges in materials science. Massive 
knowledge bases which tabulate known chemical reactions for \emph{organic} 
chemistry \cite{lawson2014making} have accelerated data-driven synthesis 
planning and related analyses \cite{segler2018planning, coley2017prediction}. 
Automated synthesis planning for organic molecules has recently achieved 
human-level planning performance using massive organic reaction knowledge bases 
as training data \cite{segler2018planning}.
There are, however, currently no comprehensive knowledge bases which systematically document 
the methods by which \textit{inorganic} materials are synthesized 
\cite{Kim2017ACS,Kim17NatureData}. Despite efforts to standardize the reporting 
of chemical and materials science data \cite{murray1999chemical},
inorganic materials synthesis procedures continue to reside as natural language
descriptions in the text of journal articles. Figure~\ref{fig-text-example} 
presents an example of such a synthesis procedure.
To achieve similar success for inorganic synthesis as has been achieved for 
organic materials, we must develop new techniques for automatically extracting 
structured representations of materials synthesis procedures from the 
unstructured narrative in scientific papers \cite{Kim17NatureData}.

To facilitate the development and evaluation of machine learning models for 
automatic extraction of materials syntheses from text, in this work we present a 
new dataset of synthesis procedures annotated with semantic structure by domain 
experts in materials science. We annotate each step in a synthesis with a 
structured frame-semantic representation, with all the steps in a synthesis 
making up a Directed Acyclic Graph (DAG). The types of nodes in the graph 
include synthesis 
operations (i.e. predicates), and the materials, conditions, apparatus and 
other entities participating in each synthesis step. Labeled edges represent 
relationships between entities, for example \emph{Condition\_of} or 
\emph{Next\_Operation}. Our dataset consists of 230 synthesis procedures 
annotated with these structures. An example sentence level annotation is given 
in Fig. \ref{fig-ann-example}. We make the corpus available to
the community to promote further research and development of scientific 
information extraction systems for procedural text.\footnote{Public dataset:
\url{https://bit.ly/2WLCbyh}}

\section{Description of the Annotated Dataset}
Here we describe the manner in which synthesis procedures 
were chosen for annotation (\S\ref{sec-selecting-syn-proc}), present a 
description of the structures we annotate 
(\S\ref{sec-structures-annotated}), summarize key statistics of the dataset 
(\S\ref{sec-dataset-stats}), highlight specific annotation decisions 
(\S\ref{sec-annotation-decisions}) and present inter-annotator agreements 
(\S\ref{sec-agreements}). All annotations were performed by three 
materials scientists using the BRAT\footnote{\protect\url{http://brat.nlplab.org/}} annotation tool \citep{stenetorp2012brat}.

\subsection{Selecting Synthesis Procedures for Annotation}
\label{sec-selecting-syn-proc}
The 230 synthesis 
procedures annotated were selected from our database of 2.5 million 
publications describing materials synthesis. The database was 
built from agreements with major scientific publication companies. Synthesis 
procedure text were obtained by first parsing the HTML text of the full 
publications, then automatically identifying candidate synthesis paragraphs 
using a trained classifier. This 
paragraph classifier was trained on a set of manually labeled paragraph examples 
and has a F1 score of 90.2 on a held out test set.\footnote{The
labeled data for this classifier is \emph{not} part of the data release 
associated with this paper due to licensing restrictions from publishers.} The 
paragraphs selected by the classifier were manually verified as containing 
complete, valid materials synthesis procedures by domain experts. While a given 
synthesis procedure is most often a single paragraph, there are cases where it 
spans multiple paragraphs, we consider all the paragraphs to be a single 
synthesis procedure. All the semantic structures were then manually 
annotated in these selected synthesis procedures. Fig \ref{fig-text-example} 
depicts an example paragraph containing a synthesis procedure. In selecting a 
synthesis procedure for 
annotation, a small number of valid synthesis procedures ($\sim20\%$) are 
ignored; this is done for the synthesis procedures which are not 
amenable to annotation from a sentence-level frame-semantic view of 
synthesis steps, or ones in which most entities in the synthesis 
do not agree with our definitions of operations and argument entities (see 
\S\ref{sec-annotation-decisions} for further discussion). 

\begin{table*}[t]
\centering
\scalebox{0.85}{
\subfloat[\label{tab-etypes}]{
    \begin{tabular}{lc}
    \textbf{Entity type} & \textbf{Count}\\
    \texttt{Material}            & 4843 \\
    \texttt{Number}             & 4095 \\
    \texttt{Operation}           & 3786 \\
    \texttt{Amount-Unit}         & 1659 \\
    \texttt{Condition-Unit}      & 1621 \\
    \texttt{Material-Descriptor} & 1430 \\
    \texttt{Condition-Misc}      & 535 \\
    \texttt{Synthesis-Apparatus} & 490 \\
    \texttt{Nonrecipe-Material}  & 475 \\
    \texttt{Brand}               & 348 
    \end{tabular}   
}
\quad
\subfloat[\label{tab-relations}]{
    \begin{tabular}{|c|l|}
    \hline
                     & \textit{Recipe-target}, \textit{Solvent-material},  \\
                                              Operation-argument  & 
\textit{Atmospheric-material}, \textit{Recipe-precursor}, \\ 
                                            relations    & 
\textit{Participant-material}, \textit{Apparatus-of} \\
                                                & \textit{Condition-of} \\ 
\hline
Non-operation entity  & \textit{Descriptor-of}, \textit{Number-of}, 
\textit{Amount-of}, \\
                             relations  & \textit{Apparatus-attr-of}, 
\textit{Brand-of}, \textit{Coref-of}, \\ \hline
Operation-Operation                    &  \multirow{ 
2}{*}{\textit{Next-operation}} 
 
\\
relations & \\
\hline                                                              
\end{tabular}
}
}
\caption{Entity types and relation labels annotated in our dataset. The table 
(a) depicts the 10 most frequent of the 21 entity types defined in our dataset, 
and the table (b) highlights the 14 relation labels among entities possible in 
our dataset.}
\label{entity-table}
\end{table*}

\subsection{Structures Annotated} 
\label{sec-structures-annotated}
An annotated graph consists of nodes denoting the participants of synthesis 
steps and edges denoting relationships between the participants in the 
synthesis. Operation nodes define the main structure of the graph and 
the arguments for each operation include different materials, conditions and 
apparatus. For annotating the text describing a synthesis procedure, we define 
a set of span-level labels that identify the operations and typed arguments in 
the text, i.e. the nodes of the graph. We also 
define a set of relationships between text spans, which label the 
edges of the synthesis graph. We detail these two kinds of labels next.

\textbf{Span-level Labels:}
Each span is a sequence of tokens or characters which form one entity 
mention (for example. ``quartz tube furnace''). Entity mentions are associated 
with 
\emph{entity types} which specify a category/kind for the entity mention. Our 
dataset defines a total of 21 entity types, with the least frequent entity 
type occuring 32 times. The 
10 most frequent entity types defined for our dataset are listed in Table 
\ref{tab-etypes}. We describe a notable subset of the entity types in more 
detail below alongside examples of their occurrence in text. In examples, the 
text underlined is the span to be annotated.
\begin{description}[noitemsep]
 \item \texttt{Material}: Materials that are used in 
the synthesis of the target. For example: \ch{Cr_2O_3}, Strontium carbonate, 
\ch{BaTiO_3}, \ch{Li_2CO_3}, Water, Ethanol.
 \item \texttt{Nonrecipe-Material}: Chemically specified materials that are 
not used in the synthesis of the synthesis target. For example: ``\ch{BaTiO_3} 
powder (\underline{\ch{Ba}}/\underline{\ch{Ti}}=0.999)'', ``\ch{Li2CO3} 
was used as the \underline{Li} source'', ``\underline{Si}/\underline{Al} ratio 
was 5''.
 \item \texttt{Operation}: Discrete actions physically performed by the 
researcher or discrete process steps taken to synthesize the target.
 \item \texttt{Material-Descriptor}: Describes a material's structure, shape, 
form, type, role, etc. and must be directly or nearly adjacent to the material 
it describes.  Does not include amounts, concentrations, or purities of 
materials. For example: \ch{CaCu_3Ti_4O_{12}} \underline{compound}, Copper 
\underline{ion}, \ch{GaAs} \underline{nanowires}, \underline{Anatase} 
\ch{TiO_2}, \underline{Deionized} water.
 \item \texttt{Meta}: A canonical name to specify a particular overall 
synthesis method used for synthesis. For example: ``Graphite 
oxide was prepared by oxidation of graphite powder according to the modified 
\underline{Hummers' method}''. ``Bi2S3 nanorods with orthorhombic structure 
were prepared through the \underline{hydrothermal method}''. ``Graphene oxide 
(GO) was prepared from graphite powder by the \underline{Staudenmaier method}.''
 \item \texttt{Amount-Unit}: These describe absolute amounts, concentrations, 
purities, ratios, flow rates and so on. For example: mg, mL, M, \%.
 \item \texttt{Condition-Unit}: These describe the units of 
measurement for intangible conditions under which operations are performed. For 
ex: \textdegree{}C, K, sec, RPM, mW.
 \item \texttt{Condition-Misc}: Qualitative descriptions of conditions. For 
example: 
Room temperature, Dropwise, Naturally, Vacuum.
 \item \texttt{Synthesis-Apparatus}: Equipment used to perform an operation 
involved in the synthesis. 
 \item \texttt{Characterization-Apparatus}: Equipment used to characterize a 
material’s properties.
\end{description}

\begin{figure*}[t]
    \centering
\subfloat[Sentences per synthesis document.]{\label{fig-syn-sent-counts}
\includegraphics[width=0.45\textwidth]{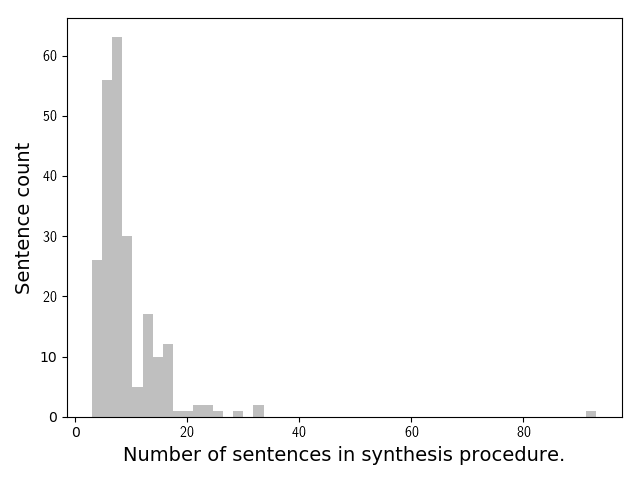}}%
    \qquad
\subfloat[Tokens per sentence.]{\label{fig-sent-tok-counts}
\includegraphics[width=0.45\textwidth]{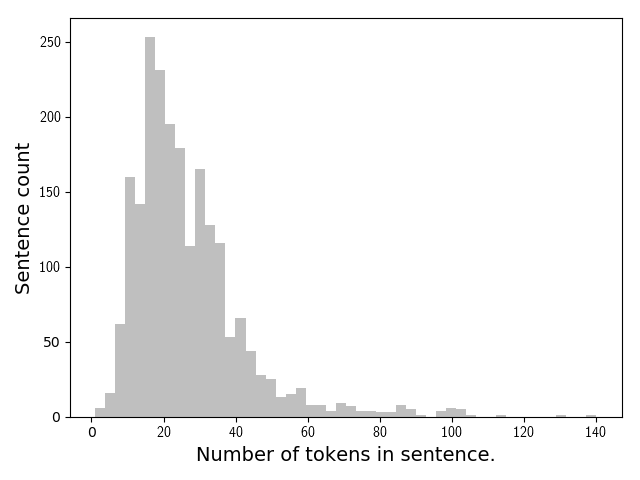}}%
    \caption{Sentences count statistics of the corpus; On average a synthesis 
procedure contains 9 sentences, each of which contain 26 tokens on average.}%
\label{fig-sent-stats}
\end{figure*}

\textbf{Relation Labels:}
We define a set of relationships between entity mentions, which label the 
edges of the synthesis graph. A subset of these relations describe direct 
relationships between operations and their arguments, others describe 
relationships between argument mentions, and the 
\textit{Next-Operation} relation describes relationships between operations so 
as to step towards annotating full recipe graphs. The different relation labels 
we define are tabulated in Table \ref{tab-relations}, a subset of these are 
defined below:
\begin{description}[noitemsep]
 \item \textit{Recipe-target}: Indicates a material assigned to an 
operation which is the target of the synthesis procedure. 
 \item \textit{Participant-material}: A material that is part of a particular 
synthesis step.
 \item \textit{Recipe-precursor}: Indicates a material which is the source of 
an element for the target material used in a specific synthesis operation.
 \item \textit{Apparatus-of}: Denotes an apparatus to be used in a synthesis 
operation.
 \item \textit{Condition-of}: Denotes a reaction condition for a synthesis 
operation.
 \item \textit{Coref-of}: Intended to capture coreferent mentions of entities 
presented by abbreviations, text in parenthesis and so on. For example: ``Air 
(\ch{O_2}/\ch{N_2} mixture gases)'' and ``He were supplied to the porous 
support 
side \dots''. ``Air'' is coreferent with \ch{O_2} and \ch{N_2}.
 \item \textit{Amount-of}: Links a number entity to the corresponding unit of 
measurement. 
 \item \textit{Next-operation}: A relation intended to denote the true 
synthesis order of the synthesis steps; the relation is also intended to 
implicitly denote the flow of intermediate materials in the synthesis. However, 
in this first release of the data, as a placeholder for future annotations, 
\textit{Next-Operation} is used simply used to indicate the next operation in 
text order rather than in true synthesis order.
\end{description}
We refer readers to our annotation guidelines for definitions of the complete 
set of entity type and relation labels in the dataset.

\subsection{Dataset Statistics}
\label{sec-dataset-stats}
Some key statistics of the dataset such as number of documents, tokens, entities 
and unique operations are listed in Table~\ref{tab:stats} and Fig. 
\ref{fig-sent-stats}.
\begin{table}[t]
  \begin{tabular}{lr}
    \bf Item & \bf Count \\ \hline
    Synthesis procedures & 230 \\
    Sentences &  2113 \\
    Tokens & 56510 \\
    Entity mentions & 20849 \\
    Entities & 4883 \\
    Unique operations & 409 \\
    Entity types (Table~\ref{tab-etypes}) & 21 \\
    Relation types (Table~\ref{tab-relations}) &  14 \\
    Avg. sentence length (Fig.~\ref{fig-sent-tok-counts}) &  26 \\
    Avg. sentences/synthesis procedure (Fig.~\ref{fig-syn-sent-counts})& 9 \\
  \end{tabular}
  \caption{Various dataset statistics. Additional details provided in referred 
figures. To determine unique operations, \texttt{Operation} entity mentions are 
lemmatized with the WordNet lemmatizer and the unique lemmas are counted.}
  \label{tab:stats}
\end{table}
In reporting these statistics we perform tokenization and sentence segmentation 
using the ChemDataExtractor package 
\cite{swain2016chemdataextractor}.\footnote{\protect
\url{https://pypi.org/project/ChemDataExtractor/1.2.2/}}

\subsection{Annotation Decisions \label{sec-annotation-decisions}}
Next we highlight specific points of contention in creating the current set of 
annotations.

\textbf{What constitutes an operation?}: While our definition of the 
\texttt{Operation} entity type specifies only actions performed by a lab 
researcher to be valid operations, there are cases where our 
annotations diverge from this definition. This happens in the following cases:
\begin{itemize}[noitemsep]
 \item Cases where an operation isn't explicitly performed by the researcher. 
For example: ``After this, the autoclave was \texttt{cooled} to room 
temperature 
naturally''.
 \item Cases with nested verb structures. For example: ``white precipitate which 
was 
\texttt{harvested} by \texttt{centrifugation} \dots''. 
\end{itemize}
In the current set of annotations, we allow experts to decide when a particular 
candidate operation should be considered valid and when it can be omitted. As 
our inter-annotator agreements will demonstrate, experts tend to agree often on 
what should be considered an operation. The question of what constitutes an 
operation is analogous to the notion of what constitutes an ``event'' in the 
broader NLP literature as highlighted by \citet{mostafazadeh2016caters}. 

\textbf{Argument state and argument re-use}: Annotation of semantic structures 
often allow for argument spans to have multiple parents 
\cite{surdeanu2008conll, banarescu2013amr, oepen2015semeval}. For example in 
Figure \ref{fig-ann-example}, the material ``Cu'' could be considered an 
argument of the operations ``placed'' and ``heated''. Allowing for arguments to 
have multiple parents however runs into complications when the operation causes 
the state of a material to change (incidentally, this is not the case 
in the example we highlight above). When a materials state changes due to a 
specific operation, considering the same text span to be the argument of a 
different operation would not be chemically valid. For example, in the sentence:
\begin{quote}
 After that, the mixed \underline{solution} was \underline{aged} at 60 degC for 
48 h, followed by \underline{heating} at 900 degC for 2 h with a heating rate 
of 5 degC min-1 in an N2 atmosphere.
\end{quote}
``solution'' is labeled as \emph{Participant-material} for ``aged'', but it 
isnt considered a \emph{Participant-material} to ``heating'' since aging caused 
it to be a different material. Similarly, in:
\begin{quote}
 1.6632 g \underline{lithium acetate} was \underline{dissolved} into 26 mL of 
\underline{ethanol-water} mixture (12:1 in volume) and slowly 
\underline{dropped} into the above \underline{suspension}.
\end{quote}
``lithium acetate'' is only labeled as \emph{Participant-material} for 
``dissolved'' and not for ``dropped'' whose sole argument is ``suspension''. 
This clearly highlights an instance of a material entirely absent from the text 
being the true argument of an operation. Therefore the current set of 
annotations does not allow for arguments to have multiple parents. Further, the 
tracking of state itself is also complicated by the difficulty in being able to 
write down precise states at a meaningful level of granularity for all possible 
materials, this is further complicated by the ambiguity presented by 
under-specified materials in synthesis text, for example in the sentence:
\begin{quote}
 With the indraught of ozone, \underline{black solid} appeared gradually and 
the \underline{clear solution} turned into \underline{black slurry} finally.
\end{quote}
Most of the entities, ``black solid'', ``clear solution'' and ``black slurry'' 
are chemically under-specified, with precise specification even unnecessary for 
describing the synthesis procedure.

\textbf{Relations across sentences}: Often, in synthesis procedures, 
a given synthesis step is described across multiple sentences. In these cases 
it would be meaningful to allow for relationships between operation-argument 
entities which are in different sentences. For the sake of simplicity and to 
stick more closely to a sentence level shallow semantic annotation, our current 
iteration of the annotations has avoided this annotation, however a very small 
number of instances of cross-sentence relations do exist ($<1\%$ of all 
relations in the dataset). Examples of this type are as follows: 
\begin{quote}
 First, sulfuric acid and nitric acid were mixed well by stirring 15 min in an 
ice bath, and then graphite powder was \underline{dispersed} into the solution. 
After 15 \underline{min}, potassium chlorate was added into the system - very 
slowly to prevent strong reaction during the oxidation process.
\end{quote}
\begin{quote}
Oxygen with 20 sccm flow rate and argon with 40 sccm flow rate were 
\underline{used} as the sputtering gas. Growth temperature was 400 
\underline{degC} and the RF power was 90 \underline{W}.
\end{quote}
Here ``min'' and ``dispersed'' are related by a \emph{Condition\_Of} relation. 
Similarly, ``degC'' and ``W'', both are annotated with \emph{Condition\_Of} 
relations to ``used''.  Annotations of this kind were created when annotators 
deemed such an annotation absolutely necessary. Synthesis procedures which 
required annotation primarily of cross-sentence relations were ignored. 

\begin{table}[t]
\centering
% \scalebox{0.85}{
\subfloat[\label{tab-typeagreements}]{
    \begin{tabular}{lc}
    \textbf{Entity type} & \textbf{Fleiss' Kappa}\\\hline
    \texttt{Material}            & 0.916 \\
    \texttt{Number}              & 0.971 \\
    \texttt{Operation}           & 0.859 \\
    \texttt{Amount-Unit}         & 0.967 \\
    \texttt{Condition-Unit}      & 0.985 \\
    \texttt{Material-Descriptor} & 0.638 \\
    \texttt{Condition-Misc}      & 0.784 \\
    \texttt{Synthesis-Apparatus} & 0.860 \\
    \texttt{Nonrecipe-Material}  & 0.371 \\
    \texttt{Brand}               & 0.862 
    \end{tabular}   
}

\subfloat[\label{tab-agreements}]{
    \begin{tabular}{lc}
\textbf{Annotation}        & \textbf{Fleiss' Kappa} \\\hline
Span-level labels & 0.861      \\
Relation labels   & 0.941
\end{tabular}
}
% }
\caption{Annotator agreements in our dataset. The table 
(a) depicts the percent agreements on 10 most frequent of the 21 entity types 
defined in our dataset, and the table (b) denotes overall agreements on the 
different annotations in our dataset.}
\label{agreement-table}
\end{table}

\subsection{Inter-annotator Agreement}
\label{sec-agreements}
Next we report a host of inter-annotator agreements for the different levels of
semantic annotation in our dataset. The agreements we report are based on a 
collection of 5 synthesis procedures which were annotated separately by all 
three expert annotators. All the numbers we report are Fleiss' Kappa scores for 
the 3 expert annotators.

\textbf{Span-level Labels:} Agreements on span level labels correspond to the 
agreement on entity type labels assigned to individual tokens. We 
observe the overall agreement on the token level labels to be 0.861. A 
break down of this agreement by the entity type is presented in 
Table \ref{tab-typeagreements}. As this indicates, there seems to be high 
agreement on labels which have clear definitions; namely.\ 
\texttt{Number}, 
\texttt{Amount-Unit}. Labels which by definition are a lot more ambiguous, 
however, have a lower agreement. The two entity 
types \texttt{Material-Descriptor} and \texttt{Nonrecipe-Material} see the 
lowest agreements. We believe these to be inherently more subjective 
entity types. In the case of \texttt{Material-Descriptor} it is often that some 
annotators may consider the descriptor and the adjacent material to be 
\texttt{Material} in its own right, for example: ``Deionized Water'' may be 
considered as a material in its own right or "deionized" may be considered to be 
a descriptor. In the case of \texttt{Nonrecipe-Material}, a similarly
harder decision needs to be made by the annotator, since these are materials 
which aren't involved in the synthesis but are still mentioned in the text for 
completeness information. Often it is up to the interpretation of the annotator
to decide whether a material is indeed involved in the synthesis leading to the 
low agreement on this entity type.

\textbf{Relation Labels:} Agreement on relation labels were computed for 
the set of cases where a pair of annotators agreed on the token 
level annotations, this happens 66\% of the time in our repeated annotations. 
For a pair of entities, if both annotators indicate the same 
relation type the annotators are considered to be in agreement. For relation 
labels we observe a agreement score of 0.941. Since we only consider cases 
where the token labels are in agreement, we believe that it is likely that when 
annotators agree on the token level annotations they also tend to agree on the 
relation level labels.

\section{Related Work}
\label{sec-related-work}
\textbf{Shallow semantic parsing in NLP:}
Prior work in the NLP community has defined and annotated semantic structures 
for text. These structured representations often seek 
to generalize about sentence level predicate-argument structure, abstracting 
away from the surface nuances of natural language and representing its 
semantics \cite{abend2017state}. A large body of work has created these 
resources for non-scientific text, as done in PropBank \cite{Palmer2005ThePB, 
surdeanu2008conll}, FrameNet \cite{fillmore2010frames}, AMR 
\cite{banarescu2013amr}, semantic dependencies \cite{oepen2015semeval} and ACE 
event schemas \cite{doddington2004automatic}. The GENIA project has defined 
event structures for biomedical data 
\cite{kim2003genia} while \citet{garg2016extracting} extended the AMR 
framework to biomedical text. Closer still to the work presented here, 
\citet{mori2014flow} have annotated cooking recipes with sentence and discourse 
level semantic relations. There has also been an interest in labeling scientific 
wetlab protocol text, with semantic 
structures and to facilitate training supervised models for the extraction of 
these structures \cite{kulkarni2018annotated}. Kulkarni et al.\ make use of 
an altered version of the EXACT2 ontology, created for the 
annotation of biomedical procedural text \cite{soldatova2014exact2}. The 
dataset presented here can be viewed to fit within 
the theme of sentence level semantics for procedural text, specifically 
tailored to materials science synthesis.

\textbf{Materials Science \& Chemistry:} Prior work in the materials science 
community  have shown that manual extraction and subsequent text mining can be 
an effective approach to analysis of synthesis routes for specific compounds 
and classes of materials \cite{raccuglia2016machine, ghadbeigi2015performance}; 
these approaches however have been limited by scale due to the manual 
extraction step. There has also been strong a consensus that comprehensively 
extracting the knowledge contained within 
written inorganic materials syntheses is a key step towards reducing the 
overall discovery and development time for novel materials 
\cite{butler2018machine}. We believe that the dataset we release fills an 
important gap in the existing work on extraction of inorganic materials 
synthesis procedures, by allowing exploration into extraction at a scale not 
attempted before. Parallel with this work, work by 
\citet{kim2018inorganic} and \citet{Tamari2019PlayingBT} adopt the dataset 
released here to aid extraction of structured representations from synthesis 
procedures and with Kim et al.\ presenting early experiments in synthesis 
planning from extracted synthesis.

The focus of existing datasets and resources in the materials 
science community, has been on materials structures and properties knowledge 
bases \cite{jain2013commentary}, rather than reactions 
and synthesis. In pursuit of more scalable methods for materials synthesis data 
extraction, \citet{young2017data} have made use of automated methods for 
extracting specific categories of materials synthesis parameters, while 
\citet{mysore2017automatically} and \citet{Kim2017ACS} have both presented 
preliminary methodological explorations for automated extraction of elements of 
a synthesis graph from materials science literature. However, these lines of 
work have not presented general purpose annotated data with which to train 
information extraction models for extraction of structured synthesis 
representations at scale, the focus of this work. 

\section{Conclusion and Future Directions}
\label{sec-conc-fut}
In this work we present a shallow semantic parsing dataset consisting of 230 
synthesis procedures. The dataset was annotated by domain experts in materials 
science. We also highlight specific difficulties in the annotation process and 
present agreement metrics on the different levels of our annotation. We believe 
the dataset will enable the development of robust supervised entity tagging 
models and is suitable for evaluating models trained to extract shallow 
semantic structures. This is evidenced by the adoption of the dataset 
by work contemporaneous with this work \cite{kim2018inorganic, 
Tamari2019PlayingBT}.

Future work in the development of this dataset could involve methods 
for the scaling up of the annotation process, perhaps by adapting the guidelines 
to enable annotation by non-experts at some stages of the annotation process. 
Further, we also plan to quantitatively establish the limits of our annotation 
schema for the kinds of information it isn't able to capture. We also plan to 
add additional layers of annotation, including: co-reference relations between 
synthesis steps, states of argument entities, and linking annotated entities to 
entries in materials science knowledge bases such as The Materials 
Project.\footnote{\protect\url{https://materialsproject.org/}} 

\section{Acknowledgements}
The authors would like to acknowledge funding from the National Science 
Foundation Award 1534340/1534341 DMREF and support from the Office of Naval 
Research (ONR) under Contract No.\ N00014-16-1-2432. Early work was 
collaborative under the Dept. of Energy’s Basic Energy Science Program through 
the Materials Project under Grant No. EDCBEE. 

\bibliographystyle{acl_natbib}
\bibliography{ms-synthesis-procedures-dataset-01}

\end{document}